\def\year{2019}\relax
\documentclass[letterpaper]{article} 
\usepackage{aaai19}  
\usepackage{times}  
\usepackage{helvet}  
\usepackage{courier}  
\usepackage{url}  
\usepackage{graphicx,tikz, multicol, epsfig}  
\usepackage{amsthm, amsmath, amssymb, mathtools}
\RequirePackage{filecontents}
\usepackage{bbm,txfonts}
\usepackage{xifthen}
\usepackage{wasysym}
\usepackage{booktabs}
\usepackage{caption}
\usepackage[margin=1.2in]{geometry} 
\usepackage[shortlabels]{enumitem}
\usepackage{todonotes}
\usetikzlibrary{shadows,shapes.arrows,calc,shapes, backgrounds, arrows,positioning}
\usepackage{xcolor}
\usepackage{mdframed}

\usepackage{hyperref}
\begin{document}
%
\title{Transfer Learning for Performance Modeling of Configurable Systems: A Causal Analysis}
\author{Mohammad Ali Javidian, Pooyan Jamshidi,  Marco Valtorta\\
Department of Computer Science and Engineering\\
University of South Carolina, Columbia, SC, USA\\
}
\maketitle
\begin{abstract}

Modern systems (e.g., deep neural networks, big data analytics, and compilers) are highly configurable, which means they expose different performance behavior under different configurations. The fundamental challenge is that one cannot simply measure all configurations due to the sheer size of the configuration space. Transfer learning has been used to reduce the measurement efforts by transferring knowledge about performance behavior of systems across environments. Previously, research has shown that statistical models are indeed transferable across environments. In this work, we investigate identifiability and transportability of causal effects and statistical relations in highly-configurable systems. Our causal analysis agrees with previous exploratory analysis~\cite{Jamshidi17} and confirms that the causal effects of configuration options can be carried over across environments with high confidence. We expect that the ability to carry over causal relations will enable effective performance analysis of highly-configurable systems.  


\end{abstract}
\section{Introduction}
To understand and predict the effect of configuration
options in configurable systems, different sampling and learning strategies have been
proposed \cite{Siegmund15,Valov17,Sarkar15}, albeit often with significant cost to cover the highly
dimensional configuration space. Recently, we performed an exploratory analysis to understand why and when transfer learning works for configurable systems~\cite{Jamshidi17}. In this paper, instead of statistical analysis, we employ \emph{causal analysis} to address the possibility of identifying influential configuration
options that have a causal relation with the performance metrics of configurable systems (identifiability) and whether such causal relations are transferable across environments (transportability).
\begin{figure}[htb]
    \centering
    \includegraphics[scale=.25]{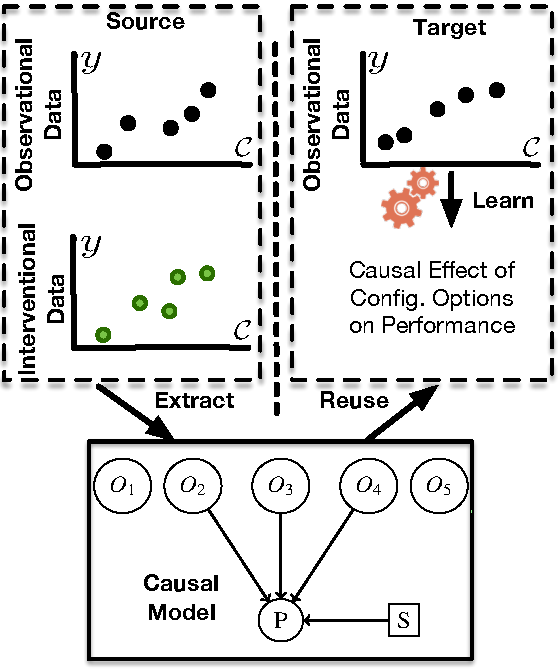}
    \caption{Exploiting causal inference for performance analysis.}
    \vspace{-0.6cm}
    \label{fig:my_label}
\end{figure}

Recently, transfer learning has been used to decrease the cost of learning by transferring knowledge about performance behavior across environments \cite{Jamshidi18,Valov17}. Fortunately, performance models typically exhibit similarities
across environments, even environments that differ substantially
in terms of hardware, workload, or version \cite{Jamshidi17}. The
challenge is to (i) identify similarities and (ii) make use of them to ease learning of performance models. 

To estimate causal effects, scientists normally perform randomized experiments
where a sample of units drawn from the population of interest is subjected to the
specified manipulation directly. In many cases, however, such a direct approach is
not possible due to expense or ethical considerations. Instead, investigators have
to rely on observational studies to infer effects. One of the fundamental questions in causal analysis is to determine when effects can be inferred from statistical information,
encoded as a joint probability distribution, obtained under normal, intervention-free measurement. Pearl and his colleagues have made major contributions in solving the problem of identifiability. Pearl \cite{Pearl95} established a calculus of interventions known as \textit{do-calculus}, consisting of–
three inference rules by which probabilistic equations involving interventions and
observations can be transformed into other such equations, thus providing a syntactic
method of deriving claims about interventions. Later, do-calculus was shown to
be complete for identifying causal effects, that is, every causal effects that can be
identified can be derived using the three do-calculus rules \cite{HuangValtorta06,ShpitserPearl06}. 

Pearl and Bareinboim \cite{BareinboimPearl11,BareinboimPearl12,pearl2014,BareinboimPearl16} provided strategies for inferring information about new populations from trial results that are more general than re-weighting. They supposed that we have available both causal information and probabilistic information for population $A$ (i.e., the source), while for population $B$ (i.e., the target) we have only (some) probabilistic information, and also that we know that certain probabilistic and causal facts are shared between the two and certain ones are not. They offered theorems describing what causal conclusions about population $B$ are thereby fixed. Conclusions about one population can be supported by information about another depends on exactly \emph{what causal and probabilistic facts they have in common}.

In this paper, we conduct a \emph{causal analysis}, comparing performance
behavior of highly-configurable systems across environmental
conditions (changing workload, hardware, and software versions), to explore when and how causal knowledge can be commonly
exploited for performance analysis. In this paper, we use the proposed formal language of causal graphs for identifiability and transportability in the literature, to answer: 

\begin{mdframed}[backgroundcolor=blue!20]
     Is it possible to identify causal relations from observational data and how generalizable are they in highly-configurable systems?
\end{mdframed}

Our results indicate the possibility of identifiability of causal effects in general. Also, our results show that many of causal/statistical relations about performance
behavior can be transferred across environments even in the most severe changes we explored, and that transportability is actually trivial for many environmental changes. Our empirical results also indicate the recoverability of conditional probabilities from selection-biased data in many cases. The results indicate that causal information can be used as a guideline for cost-efficient sampling for performance prediction of configurable systems. The supplementary materials including data and empirical results are available at: \textcolor{red}{\url{https://github.com/majavid/AAAI-WHY-2019}}.

\section{Causal Graphs}
A causal graphical
model is a special type of graphical model in which edges are interpreted as direct causal effects.
This interpretation facilitates predictions under arbitrary (unseen) interventions, and hence the
estimation of causal effects \cite{Pearl09}. In this section, we consider two constraint-based methods for estimating the causal structure from observational
data. For this purpose, we discuss the PC algorithm and the fast causal inference (FCI) algorithm~\cite{sgs}.

\subsection{Estimating causal structures}\label{estimate_causal_structure}
A causal structure without feedback loops and without hidden or selection variable can be
visualized using a directed acyclic graph (DAG) where the edges indicate direct cause-effect relationships. Under
some assumptions, Pearl \cite{Pearl09} showed that there is a link between causal
structures and graphical models. Roughly speaking, if the underlying causal structure is
a DAG, we observe data generated from this DAG and then estimate a DAG model (i.e.,
a graphical model) on this data, the estimated complete partially directed acyclic graph (CPDAG) represents the equivalence class of
the DAG model describing the causal structure. This holds if we have enough samples and
assuming that the true underlying causal structure is indeed a DAG without unobserved common causes (confounders) or selection
variables. Note that even given an infinite amount of data, we usually cannot identify the
true DAG itself, but only its equivalence class. Every DAG in this equivalence class can be
the true causal structure \cite{JSSv047i11}. 


In the case of unobserved variables, one could still visualize the underlying causal
structure with a DAG that includes all observed, unobserved cause, and unobserved selection variables. However,
when inferring the DAG from observational data, we do not know all unobserved
variables.
We, therefore, seek to find a structure that represents all conditional independence relationships
among the observed variables given the selection variables of the underlying causal structure.
It turns out that this is possible. However, the resulting object is in general not a DAG
for the following reason. Suppose, we have a DAG including observed and unobserved
variables, and we would like to visualize the conditional independencies among the observed
variables only. We could marginalize out all unbserved cause variables and condition on all unobserved selection
variables. It turns out that the resulting list of conditional independencies can in general not
be represented by a DAG, since DAGs are not closed under marginalization or conditioning \cite{RS}. A class of graphical independence models that is closed under marginalization and conditioning and that contains all DAG models is the class of \textit{ancestral graph}s \cite{RS}. A mixed graph is a graph containing three types of edges, undirected ($-$), directed ($\to$) and bidirected ($\leftrightarrow$). An ancestral graph $G$ is a mixed graph in which the following conditions hold for all vertices in $G$:
	\begin{itemize}
	    \item [(i)] if $\alpha$ and $\beta$ are joined by an edge with an arrowhead at $\alpha$, then $\alpha$ is not anterior to $\beta$.\vspace{-.2cm}
	    \item[(ii)] there are no arrowheads present at a vertex which is an endpoint of an undirected edge.
	\end{itemize}
\textit{Maximal ancestral graphs} (MAG), which we will use from now on, also obey a third rule:\vspace{-.2cm}
\begin{itemize}
    \item [(iii)] every missing edge corresponds to a conditional independence.
\end{itemize}
An equivalence class of a MAG can be uniquely represented by a \textit{partial ancestral graph}
(PAG) \cite{ZHANG2008}. 
Edge directions are marked with ``$-$" and ``$>$" if the direction is the same for
all graphs belonging to the PAG and with ``$\circ$" otherwise.
The bidirected edges come
from hidden variables, and the undirected edges come from selection variables. 


We use the  \href{https://www.hugin.com/}{Hugin PC algorithm} and the FCI algorithm in the R package \href{https://cran.r-project.org/web/packages/pcalg/index.html}{{\sf pcalg}} to recover the causal graph of each environment for our subject systems. Since all possible configurations of options are present in the first and last subject systems in Table \ref{system} and all data sets have been sampled on the basis of configuration settings alone, we can assume that there are no unobserved common causes and selection variables, i.e., the \textit{causal sufficiency assumption} \cite{sgs} holds. In other cases, due to sparsity of data, we cannot exclude the presence of hidden variables, therefore, we use the FCI algorithm to recover the causal graphs.
\section{Research Questions and Methodology}
The overall question that we explore in this paper is ``\textit{why
and when identifiability and transportability of causal effects can be exploited in configurable systems?}" We hypothesize that estimating causal effects
from observational studies alone, without performing randomized experiments or
manipulations of any kind (causal inference of this sort
is called \textit{identification} \cite{Pearl09}) is possible for configurable software systems. Also, we speculate that causal relations in the source and the target are somehow related.
To understand the notion of identification and relatedness that we
find for environmental changes, we explore three questions.
\begin{mdframed}[backgroundcolor=blue!20]
    \textbf{RQ1.} Is it  possible to estimate causal effects of configuration options on performance from observational studies alone?
\end{mdframed}
If we can establish with RQ1 that causal effects of configuration options on the performance are estimable, this would be promising for performance modeling in configurable systems because it helps us to estimate an accurate,
reliable, and less costly causal effect in an environment. Even if not all causal effects may be
estimable, we explore which configuration options are influential on performance.
\begin{mdframed}[backgroundcolor=blue!20]
    \textbf{RQ2.} Is the causal effect of configuration options on performance transportable across environments?
\end{mdframed}
RQ2 concerns transferable knowledge from source that can be exploited to learn an accurate and less costly performance model for the target environment. Specifically, we explore how the causal effects of influential options are transportable across environments and how they can be estimated.
\begin{mdframed}[backgroundcolor=blue!20]
    \textbf{RQ3.} Is it possible to recover conditional probabilities from selection-biased data to the entire population?
\end{mdframed}
RQ3 concerns transferable knowledge that can be exploited for
recovering conditional probabilities from selection-biased data to the population. 
Specifically, we explore whether causal/statistical relations between configuration options and performance measures are recoverable from biased sample without resorting to
external information.
\subsection{Methodology}
\noindent\textit{Design}: We investigate causal effects of configuration options on performance measures
across environments. So, we need to establish the performance of a system and how it is affected by configuration
options in multiple environments. As in \cite{Jamshidi17}, we measure the
performance of each system using standard benchmarks and
repeat the measurements across a large number of configurations. We then repeat this process for several changes to the
environment: using different hardware, workloads,
and versions of the system. Finally, we perform the
analysis of relatedness by comparing the performance and how
it is affected by options across environments. We perform
comparison of a total of 65 environment changes.

\noindent\textit{Analysis}: For answering the research questions, we formulate three hypotheses about:
\begin{itemize}
    \item \emph{Identifiability}: The causal effect of $X$ on $Y$ is identifiable from a causal graph $G$ if the quantity $P(y|do(x))$ can be
    computed uniquely from any positive probability of the observed variables \cite{Pearl09}.
    \item \emph{Transportability}: Given two environments, denoted $\Pi$ and $\Pi^*$, characterized by
    probability distributions $P$ and $P^*$, and causal diagrams
    $G$ and $G^*$, respectively, a causal relation $R$ is said to be
    transportable from $\Pi$ to $\Pi^*$ if $R(\Pi)$ is estimable from the
    set $I$ of interventions on $\Pi$, and $R(\Pi^*)$ is identified from
    $P, P^*, I, G$, and $G^*$ \cite{BareinboimPearl11}. 
    \item \emph{Recovering conditional probabilities}: Given a causal graph $G_s$ augmented with a
    node $S$ encoding the selection mechanism, the distribution $Q = P(y | x)$ is said to be $s$-recoverable from selection-biased data in $G_s$ if the assumptions embedded in
    the causal model renders $Q$ expressible in terms of the distribution under selection
    bias $P (v|S = 1)$ \cite{Bareinboim:2014}. 
\end{itemize}
For each hypothesis, we recover the corresponding causal graph and analyze 65 environment changes in
four subject systems mentioned below. For each hypothesis, we
discuss how commonly we identify this kind of estimation
and whether we can identify classes of changes for which
this estimation is characteristic. If we find out that for an
environmental change a hypothesis holds, it means that enough knowledge is available to estimate causal effects/ conditional probabilities across environments.

\subsection{Subject systems}
In this study, we selected four configurable software systems
from different domains, with different functionalities, and
written in different programming languages (Table \ref{system}). Further details can be found in \cite{Jamshidi17}. 
\begin{center}
\begin{table}[ht]
\caption {\footnotesize{$d$: configuration options; $\mathcal{C}$: configurations; $H$: hardware; $W$: analyzed
workload; $V$: analyzed versions.}}\label{system}
\centering
\tiny
\begin{tabular*}{\columnwidth}{@{\extracolsep{\fill}}llccccc}
\toprule
System&Domain&$d$&$|\mathcal{C}|$&$|H|$&$|W|$&$|V|$\\
\midrule
SPEAR&SAT solver&14&16384&3&4&2\\
\midrule
SQLite&Database&14&1000&2&14&2\\
\midrule
x264&Video encoder&16&4000&2&3&3\\
\midrule
XGBoost&Machine learning&12&4096&3&3&1\\
\bottomrule
\end{tabular*}
\end{table}
\end{center}

\vspace{-0.8cm}

\section{Identification of Causal Effects (RQ1)}

We can derive a complete solution to the problem of
identification whenever assumptions are expressible in a DAG form. This
entails (i) graphical and algorithmic criteria for deciding identifiability
of causal effects, (ii) automated procedures for extracting all identifiable estimand \cite{Pearl95,HuangValtorta06,ShpitserPearl06}.

Here, we investigate the possibility of estimating causal effects of configuration options on performance from observational studies alone. For this purpose, we consider a hypothesis about the possibility of identifiability in experiments with single performance metric (e.g., response time) and multiple performance metrics (e.g., response time and throughput). We expect that this hypothesis hold for (almost) all cases, which would enable an easy estimation of causal effects from the available data. 

\noindent\textbf{H1}: 
The causal effect of options $O_i$ on performance $perf$ from observed data is identifiable.

\noindent\textbf{Importance}:
If the causal effect of configuration options on performance is identifiable from available data, we can predict the performance behavior of a system in the presence/absence of a configuration option just by available observational data. Also, we may get rid of the curse of dimensionality in highly configurable systems to run and test new experiments. Because the recovered causal structure from the observed data indicates whether a given configuration option is influential on performance. 

\begin{figure}
    \centering
    \includegraphics[scale=.6]{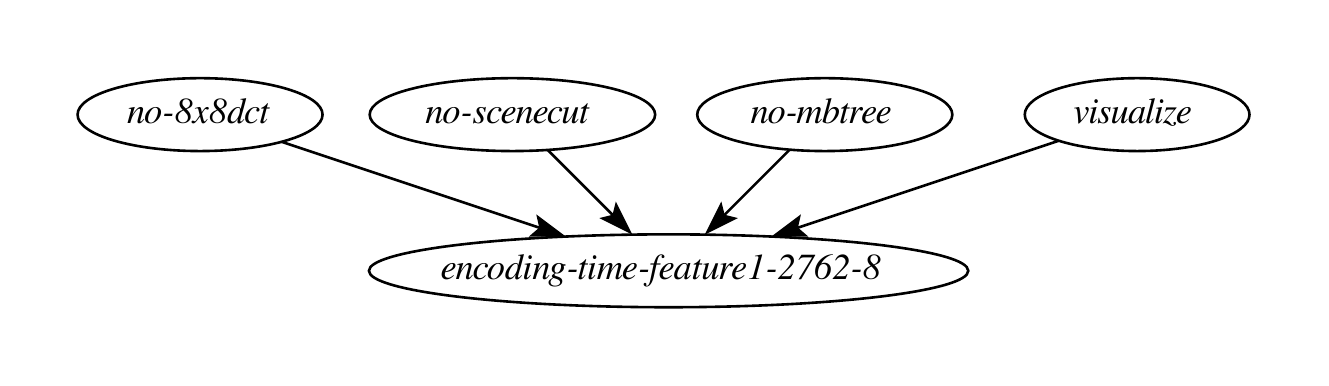}
    \caption{Causal graph for x264 deployed on internal server Feature1 and used version 2.76.2 of x264 and used a small video for encoding. For all figures we do not show options that do not affect on performance.}
    \label{fig:x264_1}
\end{figure}

\begin{figure}
    \centering
    \includegraphics[scale=.5]{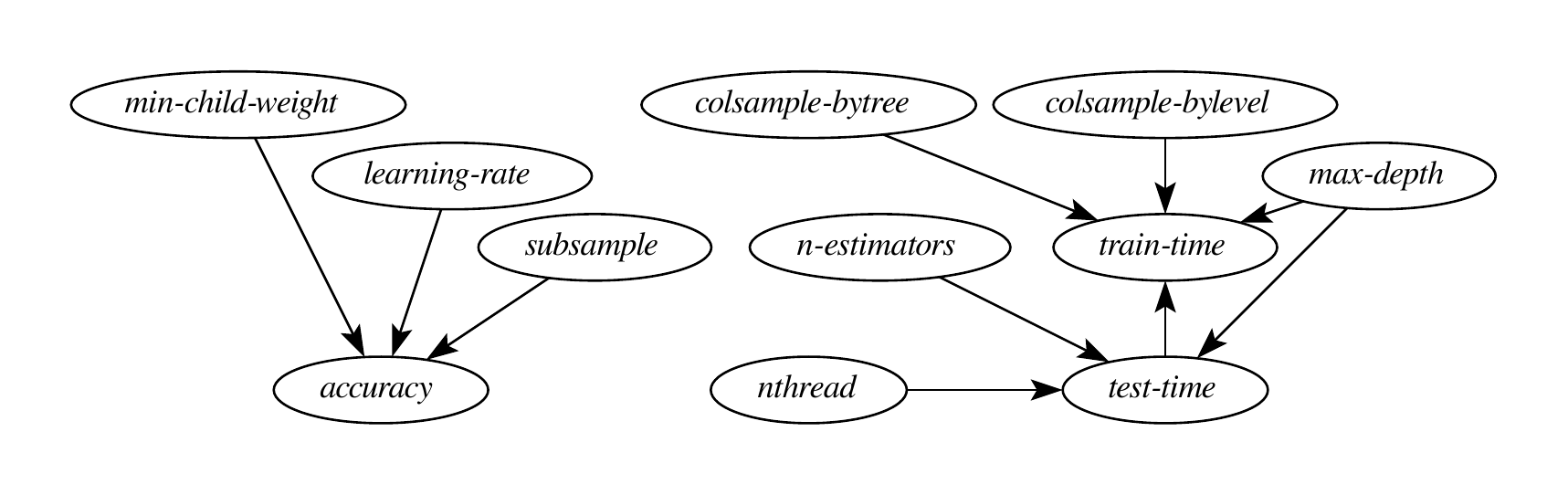}
    \caption{Causal graph for XGBoost12 with CNAE-9 data set, deployed on Feature 4. Performance nodes are: train-time, test-time, and accuracy.}
    \vspace{-0.5cm}
    \label{fig:exp-CNAE-9-data-feature4-12-1}
\end{figure}

\noindent\textbf{Methodology}: We evaluate whether $P(perf|do(O_i=o'))$ is identifiable. We used PC or FCI algorithms (with two commonly used p-values .01 and 0.05) along with
a set of background knowledge (came from experts' opinions) that explains the observed independence facts in a sample, to learn the corresponding causal graph. For example, Figure \ref{fig:x264_1} shows the obtained causal graph for x264 in the corresponding environment. We use this causal graph to estimate the causal effect of the configuration option $visualize$ on the encoding time of the system i.e.,  $P(encoding-time-feature1-2762-8|do(visualize))$. Also, Figure \ref{fig:exp-CNAE-9-data-feature4-12-1} shows  the obtained causal graph for XGBoost12 in the corresponding environment. We use this causal graph to estimate $P(test-time|do(max-depth))$.

\noindent\textbf{Results}: First, the obtained causal graph in each case indicates which configuration options are influential on performance for the corresponding environment. In all instances (see supplementary material), the number of configuration options that affect the corresponding performance metric are remarkably small (usually less than 6), indicating that the dimensionality of the configuration space for sampling and running new experiments can be reduced drastically. This observation confirms the exploratory analysis in \cite{Jamshidi17}, showing that \textit{only a small proportion of possible interactions have
an effect on performance and so are relevant}. For example, Figure \ref{fig:x264_1} shows that only four (out of 16) configuration options effect on the encoding time in the corresponding environment. Second, $P(perf|do(O_i=o'))$ is estimable in all environments with a single measurement, because in all cases, the pre-intervention and post-intervention \cite{Pearl09} causal graphs are the same, and so $P(perf|do(O_i=o'))=P(perf|O_i=o')$, indicating that the hypothesis H1 holds in general. For example, for x264 deployed on internal server Feature1 and used version 2.76.2 of x264 and used a small video for encoding, using  do-calculus and Hugin gives: $P(encoding-time-feature1-2762-8|do(visualize)=1)=P(encoding-time-feature1-2762-8|visualize=1)$ with the mean of 0.37 and a variance of  0.14.  
Also, Figure \ref{fig:exp-CNAE-9-data-feature4-12-1} shows those configuration options that affect performance nodes in the corresponding environment. Similarly, we observed  that $P(perf|do(O_i=o'))$ is estimable in all environments with multiple measurements. For example, for XGBoost12, using Rule 2 of do-calculus gives: $P(test-time|do(max-depth))= P(test-time|max-depth)$. 

\noindent\textbf{Implications}: The results indicate that such information can be used to find (causal) influential options, leading to effective exploration strategies.

\section{Transportability of Causal and Statistical Relations Across Environments (RQ2)}
Here, we investigate the possibility of transportability of causal effects across environments. For this purpose, we consider a hypothesis about the possibility of transportability of causal/statistical relations across environments. We observed that this hypothesis holds for some cases with both small and even severe environmental changes, which would enable an easy generalization (trivial transportability\footnote{This kind of transportability allows us to estimate causal/statistical relations directly from passive observations on the target environment, un-aided by causal/statistical information from the source environment \cite{BareinboimPearl11}.}) of causal and statistical relations from source to the target environment.

\noindent\textbf{H2}: The causal/statistical relation $R$ is transportable across environments. 

\noindent\textbf{Importance}: When experiments cannot be conducted in the target environment, and
despite severe differences between the two environments, it might still be possible
to compute causal relations by borrowing experimental knowledge from the source
environment. Also, if transportability is feasible, the investigator may select the essential measurements
in both experimental and observational studies, and thus minimize measurement costs.

\begin{figure}
    \centering
    \includegraphics[scale=.55]{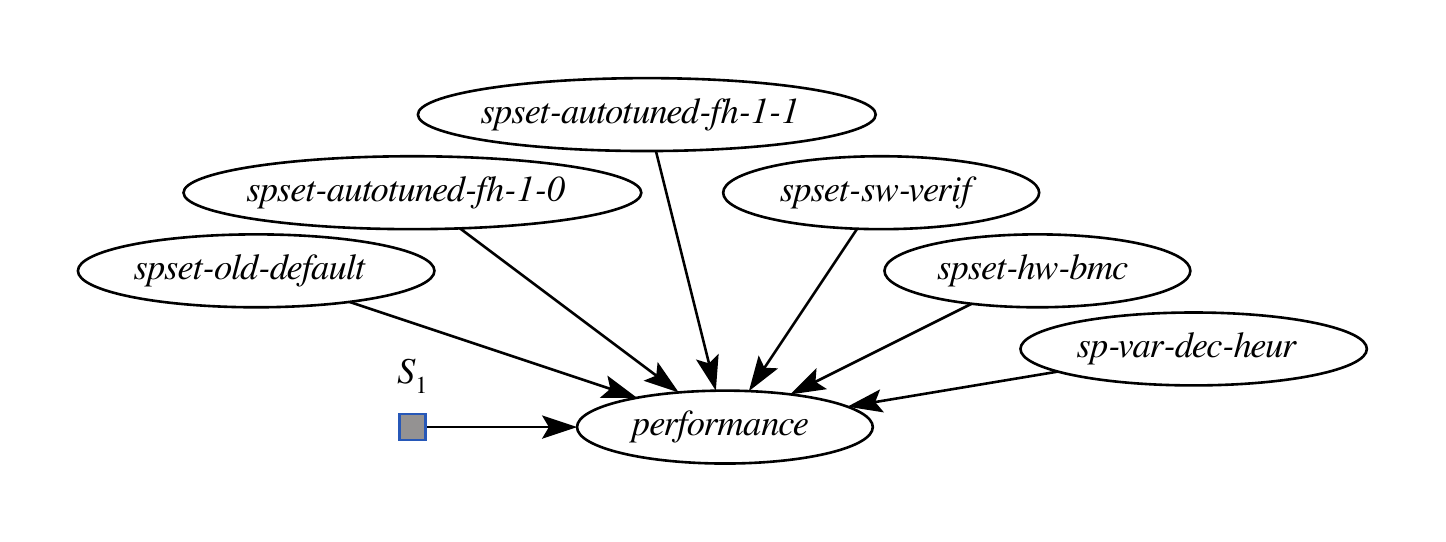}
    \caption{Selection diagram for SPEAR in two environments: one with measured solving time, deployed on a private server, version 2.7, SAT size 10286, and another deployed on Azure Cloud.}\vspace{-0.5cm} 
    \label{fig:spear}
\end{figure}

\begin{figure}
    \centering
    \includegraphics[scale=.45]{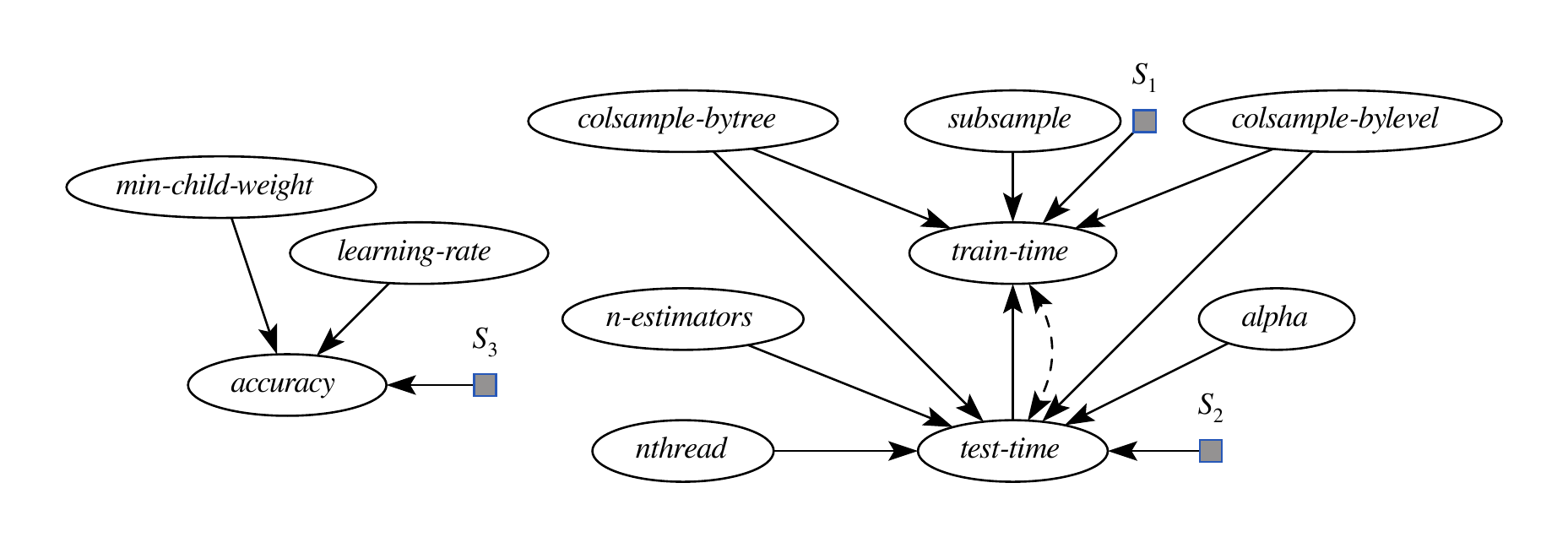}
    \caption{Selection diagram for XGBoost12 deployed on  two environments: one deployed on a private server Feature 4, with covtype dataset, and another with the same characteristics but deployed on Azure Cloud. Performance nodes are: train-time, test-time, and accuracy.}
    \vspace{-0.5cm}
    \label{fig:XGBoost12}
\end{figure}

\noindent\textbf{Methodology}: We investigate whether $P(perf|do(O_i=o'))$ (or $P(perf|O_i=o')$) is transportable across environments. For this purpose, we first recover the corresponding causal graphs for source and target environments in a similar way to that described in H1. Since the S-variables in the \textit{selection diagram}\footnote{A selection diagram is a causal diagrams augmented with a
set, S, of “selection variables,” where each member of S corresponds to a mechanism by which the two domains differ \cite{BareinboimPearl11}.} locate the mechanisms where structural discrepancies between the two environments are suspected to take place, we only add the selection node to the measurement metric node(s). For example, Figure \ref{fig:spear} shows the selection diagram for SPEAR deployed on two different environments. We use this selection diagram to verify the transportability of $P(perf|do(spset-hw-bmc))$ and $P(perf|spset-hw-bmc)$ across mentioned environments. Also, Figure \ref{fig:XGBoost12} shows  the obtained selection diagram for XGBoost12 in two environments. We use this selection diagram to verify the transportability of $P(test-time|do(colsample-bylevel))$. 

\vspace{-0.3cm}
\noindent\textbf{Results}: We observed   that H2 holds for those environments (with single measurement metric) that share the same causal graph while the presence of a selection node
pointing to the variable, say $perf$, in the selection diagram indicates that the local mechanism that assigns values to $perf$ may not the same in both environments. In these cases, the corresponding selection diagram is $O_i\to perf\gets S$, and so the causal/statistical relation is trivially
transportable \cite{BareinboimPearl11}. This observation is consistent with the exploratory analysis in \cite{Jamshidi17}, showing that for small environmental changes, the overall
performance behavior is transportable across environments. However, our observations  show
that despite glaring differences between the two environments,
it might still be possible to infer causal effects/statistical relations across environments.
Also, we observed that transportability of causal/statistical relations across environments with multiple measurement metrics. In such cases, the complete algorithm in \cite{BareinboimPearl12} can be used to derive the transport formula. Nevertheless, our observations indicate that transportable causal/statistical relations are trivial. For example, based on Figure \ref{fig:XGBoost12}, we have: $P(test-time|do(colsample-bylevel))=P(test-time|colsample-bylevel)$. 

\noindent\textbf{Implications}: {Transportability of causal relations can be exploited to avoid running new costly experiments in the target environment.}

\section{Generalizing Statistical Findings Across Sampling Conditions (RQ3)}
Here, we examine the possibility of recovering conditional probabilities from selection-biased
data. We consider a hypothesis about the possibility of recoverability without external data. We observed that this hypothesis holds for some cases, thus enabling the estimation of causal/statistical relations from selection-biased data to the entire population.

\noindent\textbf{H3}: The causal relations from selection-biased
data are transportable to the population.

\noindent\textbf{Importance}: 
Since selection bias challenges the validity of inferences in statistical analysis, we may get rid of selection bias and estimate the causal/statistical relations of entire population without resorting to
external information.

\noindent\textbf{Methodology}: We use the causal graph $G_s$
augmented with a node $S$ that encodes the selection mechanism. According to Theorem 1 in \cite{Bareinboim:2014}, the distribution $P(y|x)$ is $s$-recoverable from $G_s$ if and only if $(S\!\perp\!\!\!\perp Y|X)$, which is a powerful test for $s$-recoverability. 

\begin{figure}
\centering
    \includegraphics[scale=.5]{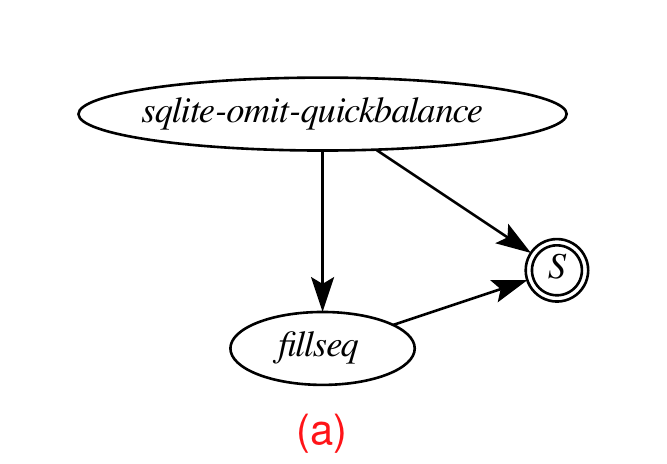}
    \includegraphics[scale=.5,page=2]{images/sqlite.pdf}
    \includegraphics[scale=.5,page=3]{images/sqlite.pdf}
    \includegraphics[scale=.5,page=4]{images/sqlite.pdf}
    \centering
    
    \caption{The causal graph $G_s$ for SQLite in the environment with Feature 20 and version 3.7.6.3.}
    \vspace{-0.6cm}
    \label{fig:sqlite}
\end{figure}

\noindent\textbf{Results}:
As we observed , in most cases, the recovered causal graph by FCI algorithm does not contain a non-chordal undirected component, indicating that FCI has not detected any selection bias from sampled data. In such cases, $s$-recoverability is the same as transportability. So, H3 holds for many cases in our study.
For example, $P(fillseq|sqlite-omit-quickbalance)$ is not $s$-recoverable in Figure \ref{fig:sqlite} (a) and (c), but it is $s$-recoverable in Figure \ref{fig:sqlite} (b) and (d).
In the data collected for the performance analysis of configurable systems, authors of \cite{Jamshidi17,Jamshidi18} sampled on the basis of configuration settings alone; therefore the conditions of Figure \ref{fig:sqlite} (b) and (d) hold, i.e., the selection bias is benign and the distribution of performance given configuration settings is recoverable. In these cases, knowledge from a sampled subpopulation can be generalized
to the entire population. 
However, FCI recovered some structures of the type of Figure \ref{fig:sqlite} (a), indicating that the sample size is small enough that some (implicit) selection bias connecting performance with one or more configuration settings. 

\noindent\textbf{Implications}: {Causal information can be used as a guideline for cost-efficient sampling for performance prediction of configurable systems and avoiding of biased estimates of causal/statistical effects in cases that recoverability was not possible.}

\section{Threats to Validity}
\textit{1) External validity}: We selected a diverse set of subject systems and a large number of purposefully selected environment changes, but, as usual, one has to be careful when generalizing to other subject systems and environment changes.

\noindent\textit{2) Internal and construct validity}: Due to the size of
configuration spaces, we could only measure configurations
exhaustively in two subject systems and had to rely on sampling (with substantial size) for the others, which may
miss causal effects in parts of the configuration space that we did not sample.

\section{Conclusion}
To the best of our knowledge, this is the first paper that exploits causal analysis to identify
the key knowledge pieces that can be exploited for transfer learning in highly configurable systems. Our empirical study demonstrate the existence of diverse forms of transferable causal effects across
environments that can contribute to learning faster, better,
reliable, and more important, less costly performance behavior analysis in configurable systems.
For a future research direction, it would be interesting to explore how
causal analysis can be employed for developing \textit{effective sampling methods} and provide \textit{explainable performance analysis} in configurable systems.

		
		
		

\subsubsection{Acknowledgments.}
This work has been supported by AFRL and DARPA (FA8750-16-2-0042).
	\bibliographystyle{aaai}\bibliography{reference}
\end{document}